\documentclass[11pt]{article}
\usepackage{acl2015}
\usepackage{times}
\usepackage{url}
\usepackage{latexsym}
\usepackage{graphicx}
\usepackage{subcaption}
\usepackage{natbib}

\title{Investigating the Impact of Text Summarization on Topic Modeling}

\author{Trishia Khandelwal \\
  {\tt trishiakhandelwal007@gmail.com}}

\date{}
\begin{document}
\maketitle
\begin{abstract}
Topic models are used to identify and group similar themes in a set of documents. Recent advancements in deep learning based neural topic models has received significant research interest. In this paper, an approach is proposed that further enhances topic modeling performance by utilizing a pre-trained large language model (LLM) to generate summaries of documents before inputting them into the topic model. Few shot prompting is used to generate summaries of different lengths to compare their impact on topic modeling. This approach is particularly effective for larger documents because it helps capture the most essential information while reducing noise and irrelevant details that could obscure the overall theme. Additionally, it is observed that datasets exhibit an optimal summary length that leads to improved topic modeling performance. The proposed method yields better topic diversity and comparable coherence values compared to previous models.
\end{abstract}

\section{Introduction}

Conventional topic models, such as Latent Dirichlet Allocation (LDA) \citep{blei2003}, Probabilistic Latent Semantic Analysis (LSA) \citep{hofmann1999}, and Non-Negative Matrix Factorization (NMF) \citep{fevotte2011}, are unsupervised approaches that have achieved significant performance in modeling documents as combinations of latent topics. These models utilize a bag-of-words representation and may need a predefined number of topics. However, the bag-of-words approach has limitations: it does not account for the semantic relationships between words, leading to weaker document representations.

Text embeddings have emerged as a promising alternative to the bag-of-words approach for creating document representations. Top2Vec \citep{angelov2020} finds topic vectors using joint document and word semantic embeddings. The Bidirectional Encoder Representations from Transformers (BERT) \citep{devlin2018} model, a large language model (LLM), has demonstrated strong capability in producing word and sentence embeddings that effectively capture contextual information. LLM-based embeddings convert documents into high-dimensional vectors, where closer points in the vector space represent similar texts, which are then grouped together.

BERTopic \citep{grootendorst2022} uses a pre-trained language model to generate document-level embeddings. The dimensionality of these embeddings is then reduced to optimize the clustering process. A custom class-based variation of TF-IDF is subsequently used to extract the topic representations. BERTopic has shown strong performance on benchmark datasets and serves as a base model for this study.

This paper proposes a method to improve topic modeling performance by preprocessing documents before inputting them into the BERTopic model. To address domain-specific stopwords and other noise in the documents, an LLM with few-shot learning prompts has been proposed to generate summaries that capture the core essence of the content. These summaries are then used as input into the BERTopic model, resulting in improved topic modeling. This paper examines how variations in input data affect topic modeling outcomes by comparing the topic model's performance on summaries of different lengths. The concept of an optimal summary length which yields the best performance for a given dataset is introduced. Additionally, methods for improving the robustness of topics generated through topic modeling are explored. The aim is to inspire further research into the application of deep learning for topic modeling by providing a detailed methodology that enhances the performance of neural topic models.

\section{Related Work}

\subsection{Topic Models}
\label{sect:pdf}

The Combined Topic Model (CTM) uses contextual embeddings from a pre-trained model as input into a neural topic model \citep{bianchi2021}. DeTiME leverages encoder-decoder LLMs to generate highly clusterable embeddings. It enhances clusterability and semantic coherence by adapting the LLM architecture \citep{xu2023a}. BERTopic uses sentence embeddings, dimensionality reduction, and clustering techniques like HDBSCAN to generate topics \citep{grootendorst2022}. Even though semantics and context is taken into account during the creation of document embeddings, it needs to be improved in the reconstruction process at the end. Thus, having more directed and concise input data will enhance the topic reconstruction at the end which depends on the bag of words approach.

Some recent studies have explored the direct use of LLMs in topic modeling.  PromptTopic employed models like GPT-4 \citep{openai2023b} and LLaMA \citep{touvron2023} to generate broad document categories, which were subsequently refined into topic words by the model being used \citep{wang2023}. Similarly, \citep{mu2024} utilized these models for basic prompting, both with and without seed topics, followed by LLM-driven topic summarization and merging.

\subsection{Language Models}

Transformer-based models such as BERT \citep{devlin2018}, GPT-3 \citep{brown2020},  and GPT-4 \citep{openai2023b} demonstrate exceptional performance across a wide range of language tasks. Their self-attention mechanisms enable effective handling of long-range dependencies. The GPT-3.5 Turbo models \citep{openai2023a} are specifically optimized for chat applications, enhancing their capability to understand and generate natural language. The FlanT5 model \citep{chung2022}, an encoder-decoder architecture, has undergone instruction fine-tuning on a variety of tasks, resulting in high performance across many language tasks.

\section{Methodology}

The proposed methodology consists of two main components: summarization and topic modeling, as illustrated in Figure 1.

\begin{figure*}
    \centering
    \includegraphics[width=1\linewidth]{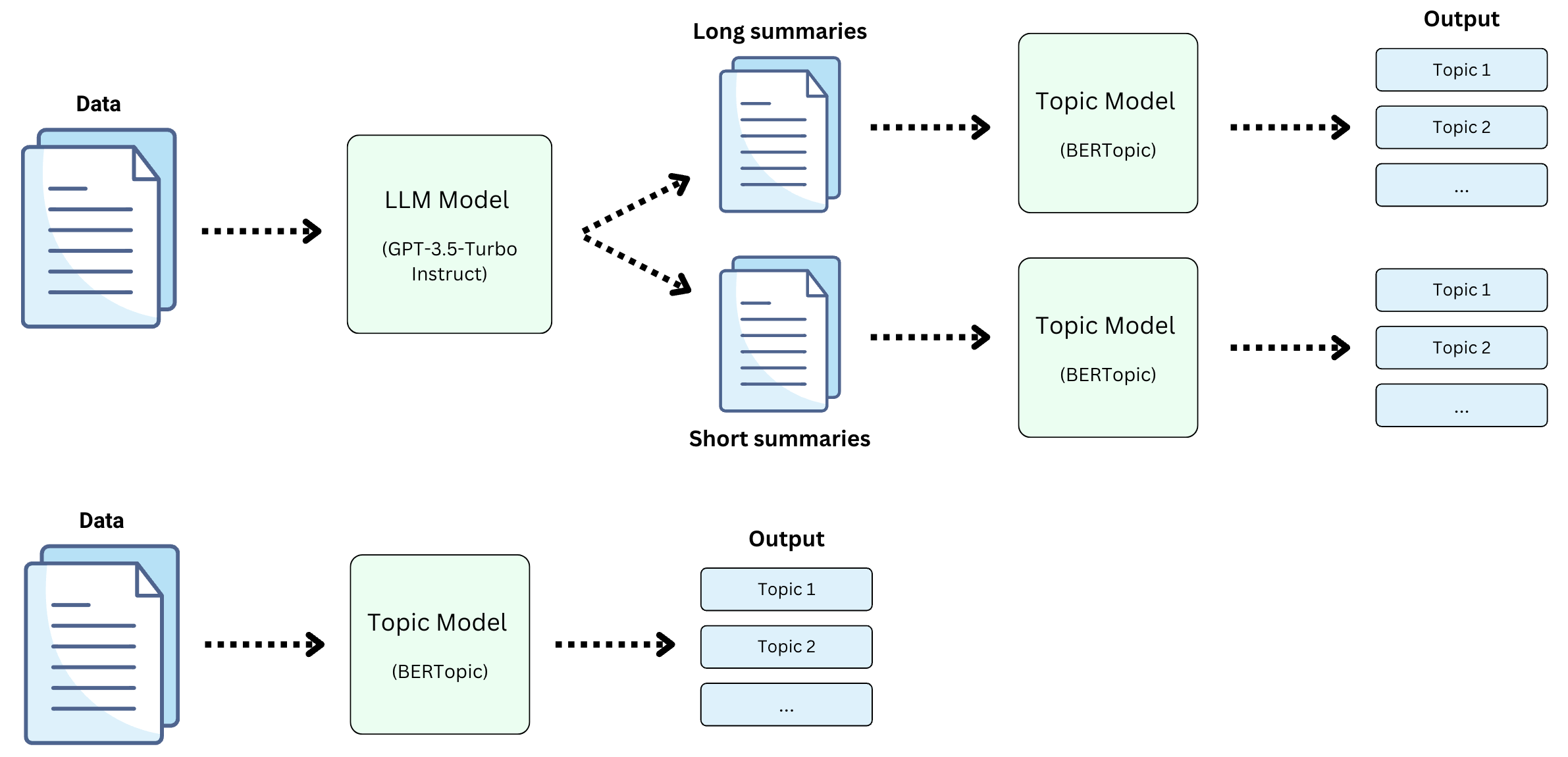}
    \caption{This diagram outlines three experimental cases: two where LLM-generated summaries are used as inputs for the topic model, and one where the original data is used directly for topic modeling.}
    \label{fig:enter-label}
\end{figure*}

\subsection{Summarization}

Document summarization is employed to reduce noise in the data. To determine the optimal length for the summaries, two sets of summaries were generated: one with 20-30 words for concise and pointed summaries and another with 60-80 words for more contextual and detailed summaries. The length criteria were chosen to balance brevity and context, with the shorter summaries conveying the main keywords of the document, and the longer summaries providing a deeper insight into its themes by covering the main points. 

For summarization, the GPT-3.5-Turbo-Instruct model \citep{openai2023a} was used for its ability to interpret and execute instructions proficiently and accurately. Designed for use with the completions endpoint, this model requires an open-ended prompt with context, providing increased flexibility and control over the generated results. 

To generate robust and consistent summaries, a few-shot prompting technique was employed. This involved providing the model with two sample documents along with their summaries. In addition, a task description that utilized the persona pattern was provided, instructing the model to assume the role of a summarization expert. 

The GPT-3.5-Turbo-Instruct model has a maximum context window size of 4096 tokens. This issue was addressed by truncating all input documents to a fixed word count. This was necessary to accommodate the larger documents within the model’s input constraints. This truncation affected less than 1\% of the documents across the three datasets, minimizing any potential impact on the overall quality of the summaries.

\subsection{Topic Modeling}

BERTopic \citep{grootendorst2022} is the topic model utilized in our experiments. BERTopic first generates document embeddings using the Sentence-BERT (SBERT) framework \citep{reimers2019}. The dimensionality of these embeddings is then reduced using UMAP \citep{mcinnes2018}. The reduced embeddings are subsequently clustered with HDBSCAN \citep{mcinnes2017}. To form topic representations, BERTopic employs a modified version of TF-IDF known as class-based TF-IDF (c-tf-idf). Maximal Marginal Relevance (MMR) \citep{carbonell1998} is used to enhance the diversity of topic representations. In the experiments, various MMR values (0.1, 0.2, 0.3) and different minimum documents per topic values (depending on the dataset size) were tested across the different input types: full text, short summaries, and long summaries.

\begin{table*}
    \centering    \begin{tabular}{|c|c|c|c|c|} \hline   
         & \multicolumn{2}{|c|}{\textbf{BBC}} & \multicolumn{2}{|c|}{\textbf{20 Newsgroup}}\\ \hline
 & Diversity& Coherence& Diversity& Coherence\\\hline
 LDA& 0.729&0.491&0.827&0.507\\\hline
 Top2Vec& 0.743& 0.377& 0.619& 0.469\\ \hline  
 NMF& 0.96& 0.545& 0.96&0.620\\\hline
 BERTopic& 0.940& \textbf{0.694} & 0.959&\textbf{0.676}\\ \hline 
 \multicolumn{5}{|c|}{}\\\hline
 Short Summary + BERTopic&  0.947&  0.693 & \textbf{0.988}&0.664\\ \hline  
 Long Summary + BERTopic&  \textbf{0.974}&  0.664 & 0.957&0.536\\\hline 
    \end{tabular}
    \caption{The diversity and coherence values obtained by various topic modeling techniques and our method on the BBC  and 20 Newsgroups datasets.}
    \label{tab:my_label}
\end{table*}

\section{Experimental Setup}

\subsection{Datasets}

Two datasets are used to validate the performance of our method: BBC News and 20 NewsGroups. 

The BBC News dataset contains 2,225 documents across five main areas from the BBC News website, spanning from 2004 to 2005 \citep{greene2006}. The dataset is accessed from Kaggle. The 20 NewsGroups dataset contains around 18,000 documents spanning across 20 categories \citep{lang1995}. The Scikit-Learn library is used to access the dataset. These datasets were chosen for their diversity and variation in terms of document length and content.

\subsection{Evaluation}

The effectiveness of the approach is evaluated using two commonly used metrics: topic coherence and topic diversity. Coherence measures the interpretability of a topic by humans, focusing on the semantic similarity of the words representing a particular topic. Topic coherence is assessed using C\textsubscript{V} coherence, which combines indirect cosine similarity and normalized pointwise mutual information (NPMI) \citep{bouma2009} to calculate pairwise word similarity within a sliding window \citep{roder2015}. C\textsubscript{V} typically ranges from 0 to 1, with higher values indicating greater coherence. Topic diversity, on the other hand, measures the percentage of unique words across all topic representations. This metric ranges from 0 to 1, where higher scores represent more varied topics.

The performance of the proposed method is compared against regular BERTopic applied on the full text, LDA, and vONTSS \citep{xu2023b}. The indicated performance of LDA is the average of three runs with a number number of topics. These numbers are chosen in intervals of five around the number with the best performance. vONTSS was run using the author’s code publicly available on GitHub and the results recorded are the average of three runs. 

\subsection{Experimental Settings}

To ensure robustness in the results, the topic model is run three times for each combination. The average for the metrics is taken and recorded. Summaries are generated once, and experiments are conducted on both datasets. Three MMR values, three minimum documents per topic values, and three input types (full text, short summaries, and long summaries) are tested, leading to a total of 54 combinations.  For each topic, the top ten keywords are extracted to represent it, and these keywords are subsequently used to calculate the metrics for that run. Although the input to the topic model may be summaries, the metrics are computed using the topic keywords relative to the entire dataset to allow for a uniform comparison with other methods.

\begin{figure*}[ht]
    \centering
    \subcaptionbox{Diversity of the topics generated by BERTopic given the original BBC News dataset and the two types of summaries.}
        {\includegraphics[width=0.48\linewidth]{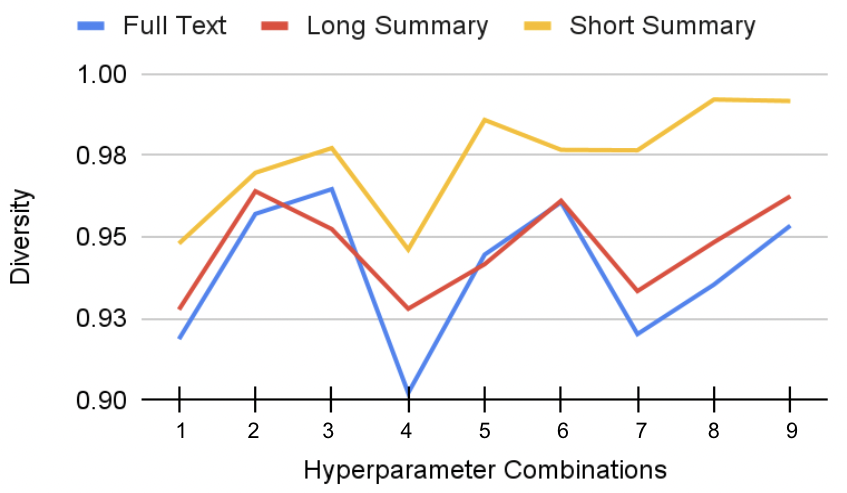}}
    \hspace{1em}
    \subcaptionbox{Coherence of the topics generated by BERTopic given the original BBC News dataset and the two types of summaries.}
        {\includegraphics[width=0.48\linewidth]{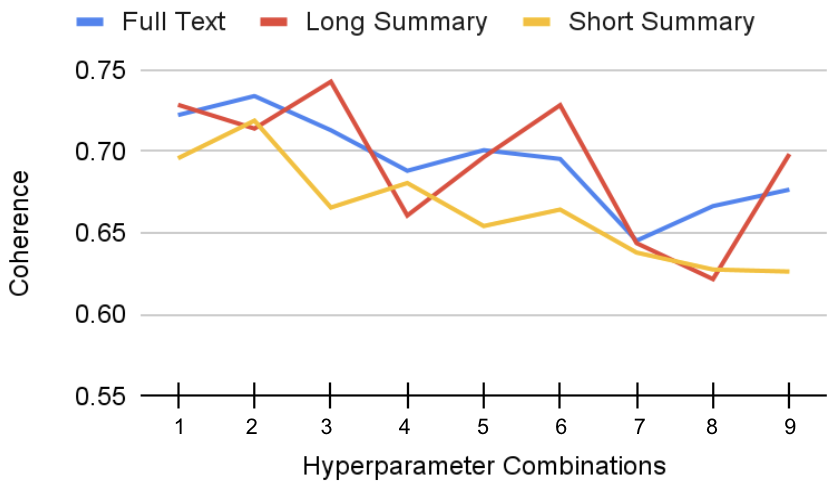}}
        \\[2em]
    \subcaptionbox{Diversity of the topics generated by BERTopic given the original 20 Newsgroup dataset and the two types of summaries.}
        {\includegraphics[width=0.48\linewidth]{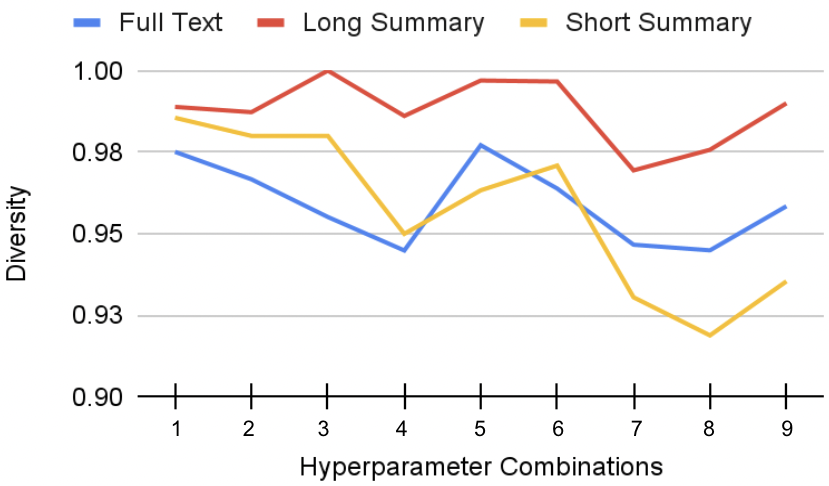}}
    \hspace{1em}
    \subcaptionbox{Coherence of the topics generated by BERTopic given the original 20 Newsgroup dataset and the two types of summaries.}
        {\includegraphics[width=0.48\linewidth]{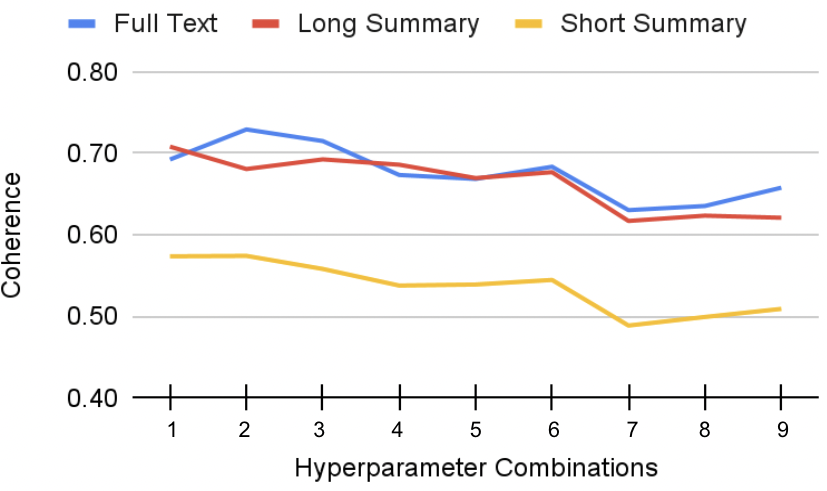}}
    \caption{Topic modeling performance for various hyperparameter combinations.}
    \label{fig:grouped-figures}
\end{figure*}

\section{Results and Discussion}
Varying the MMR values and the minimum topic size did not significantly impact the results. Figure 2 illustrates these outcomes. Each graph shows the diversity or coherence metrics for one of the datasets, depicted through a multi-line chart. The charts compare the results obtained by applying BERTopic to the entire text, short summaries, and long summaries. The lines on the chart connect points that represent the average results from three trials conducted for each of the nine combinations of MMR values and minimum topic sizes tested per dataset and input type. Each point on the graph corresponds to one of these combinations, providing a visual comparison of how the different input types—full text, long summaries, and short summaries—perform under varying conditions. The overall average of these nine values for each input type is summarized in Table 1. 

The diversity values produced using either of the summaries consistently outperform those obtained using the entire text, as can be seen in Figures 2(a) and 2(c). The BBC dataset comprises longer documents, with an average length of approximately 380 words per document. It shows better results when shorter summaries are used. In contrast, the 20 Newsgroups dataset, which contains shorter documents averaging 180 words, performs better with longer summaries. These results suggest that the length of the summaries plays a critical role in the effectiveness of summarization for topic modeling. Longer documents tend to contain more diverse information, so shorter summaries can effectively distill the main points and focus on core content, helping to highlight primary themes. In contrast, shorter documents may lack sufficient context, which is why longer summaries will help preserve subtleties and background information, providing the topic model with more, yet still focused, information to work with.

The results suggest that an optimal summary length may exist for each dataset, depending on factors such as the original document length, content density, and quality. Experimentation with various summary lengths on a sample of a dataset tailors the approach to the specific attributes of the data and will give better topic modeling results. However, it’s important to note that summarization did not consistently result in significantly higher coherence values, as seen in Figures 2(b) and 2(d). The coherence values obtained from long summaries are comparable to those from the full text, while short summaries performed worse. This lower performance of short summaries may be due to the condensation of multiple separate ideas into fewer words. Since coherence is measured based on the proximity of words within the entire document, the coherence among the words in the summaries tends to be lower, which in turn causes the coherence of the top 10 words in the topic to also decrease.

\section{Limitations}

There are several limitations to the proposed approach. First, the effectiveness of the proposed method heavily relies on the quality of the summaries generated by the LLM in use. In this study, GPT-3.5-Turbo-Instruct \citep{openai2023a} was employed, which is a closed-source model, limiting our control and insight into its inner workings. Additionally, the model's context window size poses a limitation, as it may lead to potential information loss. Although this was not a significant issue in this study, it could become problematic for datasets with larger documents. The proposed approach is also not suitable for datasets with very short documents, such as tweets, where summarization is neither feasible nor effective. Attempting to summarize such brief texts may lead to worse results.

\section{Conclusions and Future Work}

A method is proposed to enhance topic modeling performance by leveraging a large language model to generate document summaries, which are subsequently used as inputs to BERTopic. The experiments with different summary lengths provide insights into the optimal summarization length that balances noise reduction and preservation of each document's essence. Based on the findings, it is recommended to test a range of summary lengths on a sample of the dataset to determine the best approach for topic modeling, as the optimal length may vary depending on the dataset.

Further research is needed to further improve topic modeling and to address the weaknesses of this study. A key priority should be enhancing topic coherence, which could be achieved by refining the summarization process or incorporating coherence-focused optimizations during the topic modeling phase. To evaluate the broader applicability of the proposed approach, it should be tested on a more diverse set of datasets, particularly those beyond news documents. Given that the proposed method relies heavily on the quality of LLM-generated summaries, advancements in LLM technology are expected to enhance topic modeling outcomes. Future studies could also explore the impact of using alternative LLMs to further assess and optimize topic modeling performance.

\section{Acknowledgments}

I would like to express my gratitude to Advay Pal for his mentorship and continuous guidance throughout my research journey.

I also extend my sincere thanks to Dr. Manoj Khandelwal for his advice and thorough review of the manuscript, which greatly improved its quality.

\bibliographystyle{acl_natbib}
\bibliography{custom}

\end{document}